\documentclass[a4paper]{spie}  

\usepackage{amsmath,amsfonts,amssymb}
\usepackage{graphicx}
\usepackage{subcaption}
\usepackage[colorlinks=true, allcolors=blue]{hyperref}

\title{CycleGAN without checkerboard artifacts for counter-forensics of fake-image detection}

\author[a]{Takayuki Osakabe}
\author[b]{Miki Tanaka}
\author[c]{Yuma Kinoshita}
\author[d]{Hitoshi Kiya}
\affil[a-d]{Tokyo Metropolitan University, Tokyo, Japan}

\authorinfo{Further author information: \\T. Osakabe: E-mail: osakabe-takayuki@ed.tmu.ac.jp\\  M. Tanaka: E-mail: tanaka-miki@ed.tmu.ac.jp\\ Y. Kinoshita: E-mail: ykinoshita@tmu.ac.jp\\ H. Kiya: E-mail: kiya@tmu.ac.jp}

\begin{document} 
\maketitle

\begin{abstract}
In this paper, we propose a novel CycleGAN without checkerboard artifacts
for counter-forensics of fake-image detection.
Recent rapid advances in image manipulation tools and 
deep image synthesis techniques, such as Generative Adversarial Networks (GANs)
have easily generated fake images, so detecting manipulated images 
has become an urgent issue. 
Most state-of-the-art forgery detection methods assume that images include
checkerboard artifacts which are generated by using DNNs.
Accordingly, we propose a novel CycleGAN without any checkerboard artifacts 
for counter-forensics of fake-mage detection methods for the first time,
as an example of GANs without checkerboard artifacts.
\end{abstract}

\keywords{GAN, checkerboard artifacts, counter-forensics, deep learning}

\section{INTRODUCTION}
\label{sec:intro}  
Although deep neural networks (DNNs) have led to major breakthroughs 
in computer vision, for a wide range of applications, 
they have created new concerns and problems. 
DNNs in general suffer from attacks such as invasion of 
data privacy\cite{Warit_1, Warit_2, Warit_3},
and adversarial attacks\cite{Maung_1, Maung_2, Maung_3}. 
In addition, recent rapid advances in deep image synthesis techniques,
such as Generative Adversarial Networks (GANs)\cite{GAN} 
have easily generated fake images,
so detecting manipulated images has become an urgent issue.

So far, a lot of researchers have investigated forgery detection methods,
in which most state-of-the-art forgery detection methods assume that
images include checkerboard artifacts which are generated 
by using DNNs\cite{AutoGAN, CNNdetect}.
In contrast, checkerboard artifacts-free DNNs have been proposed by using a
fixed convolutional layer\cite{Kinoshita, Sugawara_1, Sugawara_2}, 
but the technique for avoiding the artifacts
have never been applied to img2img GANs like CycleGAN.
In this paper, we propose a CycleGAN without checkerboard artifacts for 
the first time, and moreover, the proposed CycleGAN is demonstrated to be
effective for counter-forensics of fake image detection methods, 
as an example of GANs without checkerboard artifacts.

\section{RELATED WORK}
\label{sec:related}

\begin{figure}[htb]
 \begin{minipage}{0.32\columnwidth}
  \centering
   \includegraphics[width=50mm]{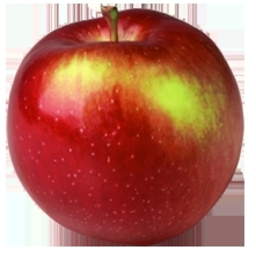}
  \subcaption{real image}
  \label{real}
 \end{minipage} 
 \begin{minipage}{0.32\columnwidth}
  \centering
   \includegraphics[width=50mm]{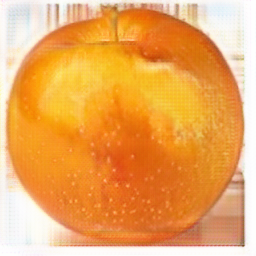}
  \subcaption{fake image with artifacts}
  \label{fake}
 \end{minipage} 
  \begin{minipage}{0.32\columnwidth}
  \centering
   \includegraphics[width=50mm]{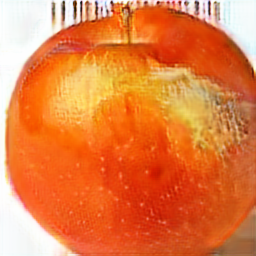}
  \subcaption{fake image without artifacts}
  \label{fixed}
 \end{minipage}
 \caption{Example of fake images}
 \label{Fig1}
\end{figure}

\begin{figure}[htb]
\centering
\includegraphics[width=120mm]{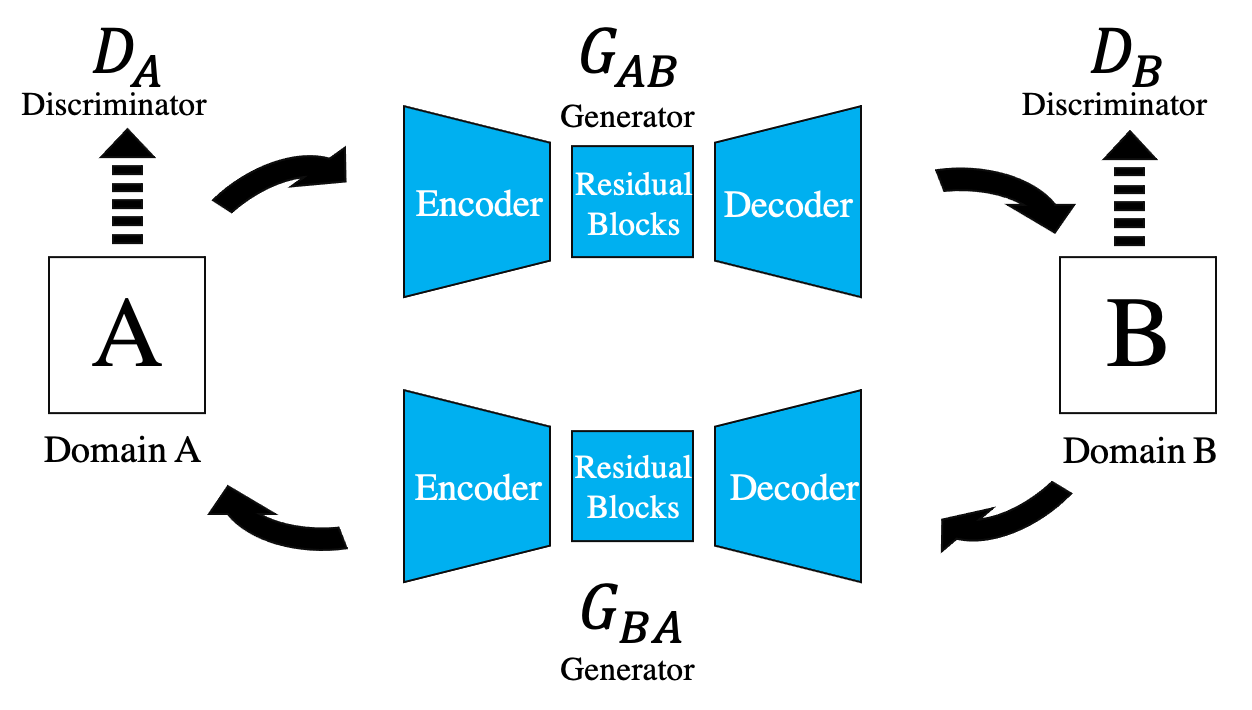}
\caption{CycleGAN model}
\label{fig:cyclegan}
\end{figure}

CycleGAN allows us to convert one image into another,
called image to image conversion.
Figure \ref{Fig1} shows an example of images generated by using CycleGAN.
CycleGAN is a model for unsupervised learning of image to image translation
using an adversarial network with cycle-consistency, 
as shown in Fig.\ref{fig:cyclegan}. 
As shown in the figure, CycleGAN has two generators: $G_{AB}$ that converts 
images from domain A to domain B, and vice versa, $G_{BA}$. 
Discriminator $D_B$ encourages $G_{AB}$ to translate A into outputs
indistinguishable from domain B. 
In this paper, we refer to images generated by GANs as fake images. 
Many researchers have been working on detecting images generated 
by using GANs\cite{AutoGAN, CNNdetect}.
In typical fake-image detection methods such as 
the method proposed by Zhang et al.\cite{AutoGAN}, 
fake-images are detected by finding checkerboard artifacts 
included in manipulated images.

\section{CYCLE-GAN WITHOUT CHECKERBOARD ARTIFACTS}
\label{sec:cgan}

\begin{figure}[htb]
\centering
\includegraphics[width=140mm]{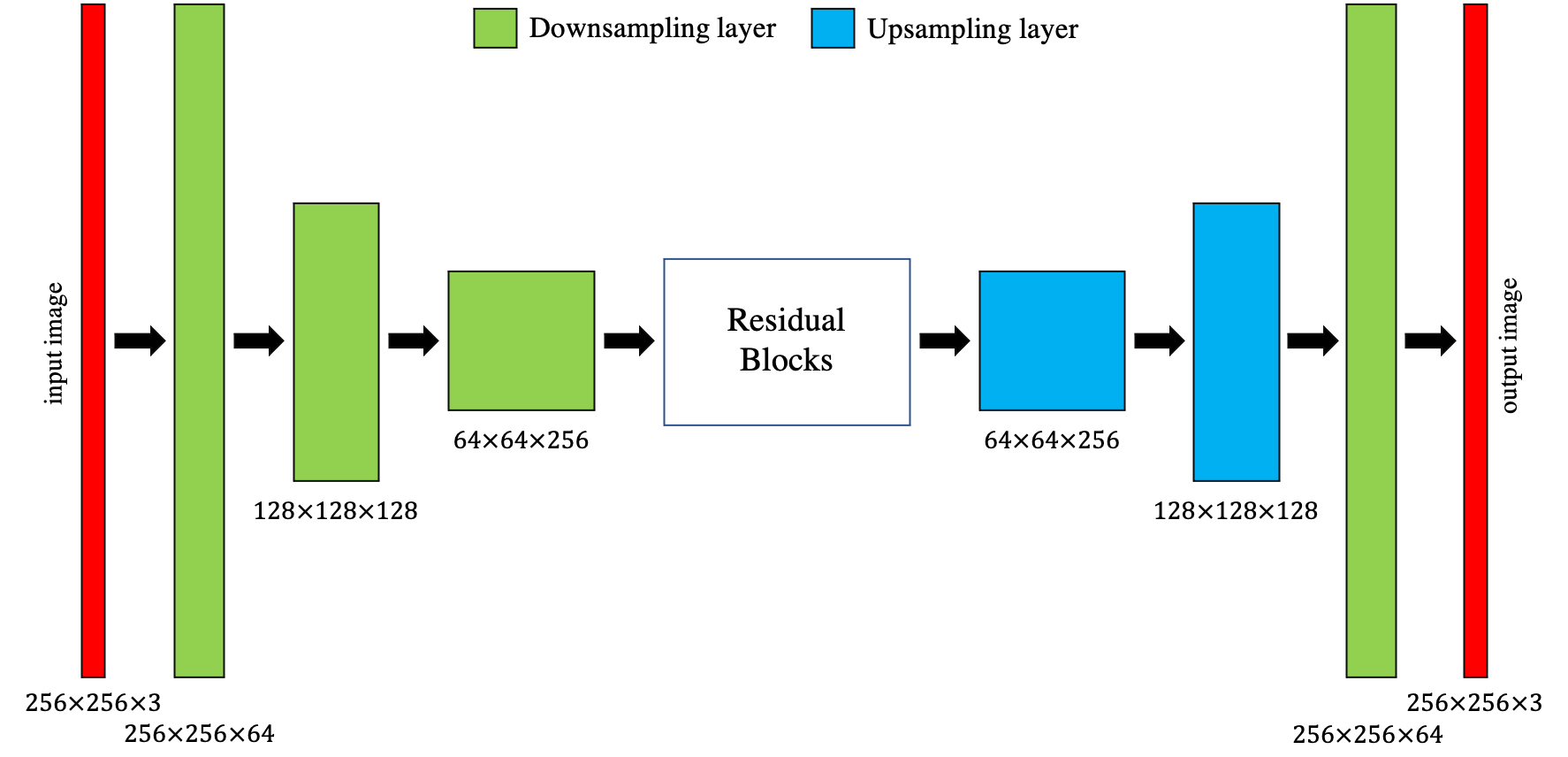}
\caption{Generator of CycleGAN. Each box denotes multi-channel feature map produced by each layer. Output size of each layer is denoted under each box($height \times width \times channel$).}
\label{gen}
\end{figure}

\begin{figure}[htb]
\centering
\includegraphics[width=150mm]{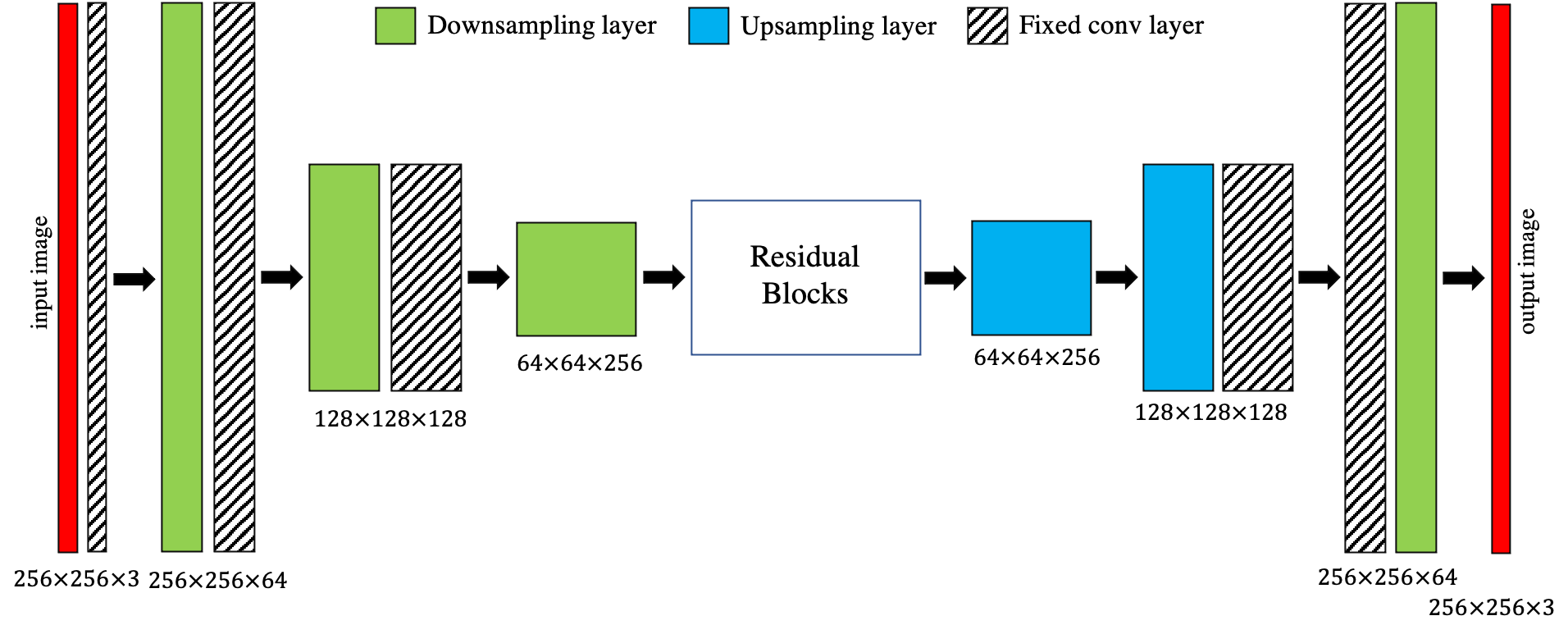}
\caption{Generator of proposed CycleGAN without checkerboard artifacts. Each box denotes multi-channel feature map produced by each layer. Output size of each layer is denoted under each box($height \times width \times channel$).}
\label{gen_p}
\end{figure}

Checkerboard artifacts are known to be caused in upsampling layers during
forward propagation and in the downsampling layer during back propagation as
in\cite{Kinoshita, Sugawara_1, Sugawara_2}. 
By inserting a fixed convolution layer into upsampling and
downsampling layers, the artifacts were demonstrated to be completely avoided,
although this technique has not been applied to any image to 
image transformations like CyclcGAN.
We have two aims in this paper. 
The first one is to apply this technique to CycleGAN to remove 
checkerboard artifacts included in fake images.   
The another is to confirm the effectiveness in counter-forensics against 
fake-image detection. 

The structure of the generator of CycleGAN is shown 
in the Fig.\ref{gen}\cite{CycleGAN}. 
In contrast, Fig.\ref{gen_p} shows the structure of 
the generator of the proposed CycleGAN without checkerboard artifacts, 
where fixed convolution layers are inserted in the encoder and decoder sections.
In this paper, a fixed convolution layer is inserted into 
both every upsampling layer and every downsampling layer
as in the paper\cite{Kinoshita}.
To constrain the Lipschitz constant, 
we also inserted Spectral Normalization layer\cite{SpectralNorm} into 
the discriminator and changed the Adversarial loss to Hinge loss.

\section{EXPERIMENT}
\label{sec:experiment}

\begin{figure}[htb]
\begin{minipage}{0.32\columnwidth}
 \centering
  \includegraphics[width=40mm]{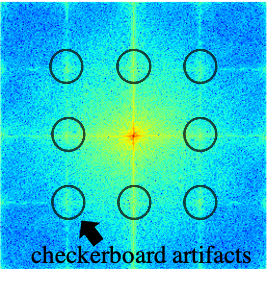}
  \subcaption{Image in Fig.\ref{Fig1}(b)}
  \label{2Dspec_1}
\end{minipage}
\begin{minipage}{0.32\columnwidth}
 \centering
  \includegraphics[width=40mm]{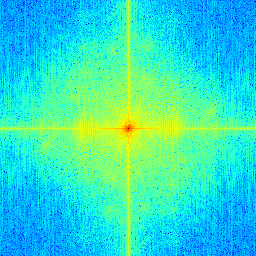}
  \subcaption{Image in Fig.\ref{Fig1}(c)}
  \label{2Dspec_2}
\end{minipage}
\begin{minipage}{0.32\columnwidth}
 \centering
 \includegraphics[width=60mm]{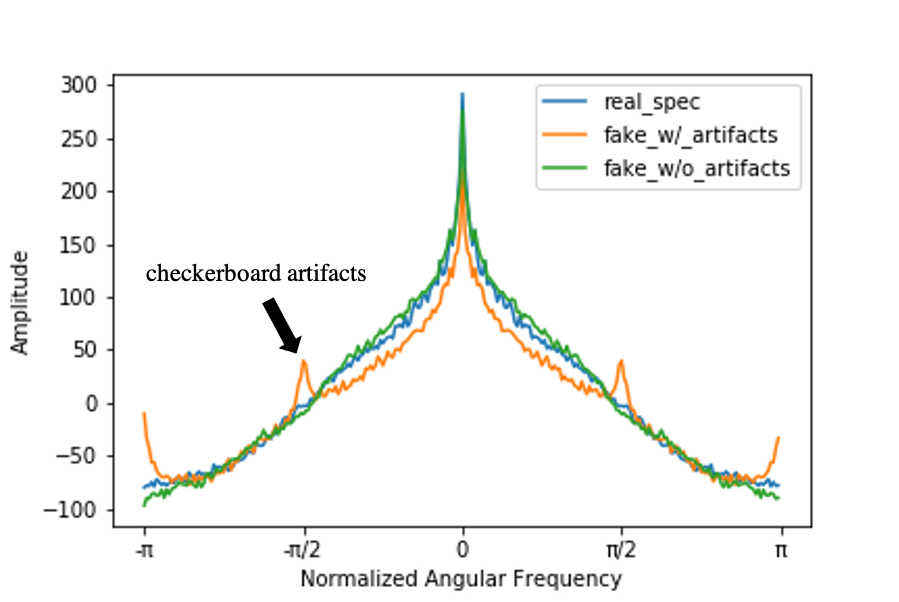}
 \subcaption{Images in Fig.\ref{Fig1}(1D)}
 \label{1Dspec}	
\end{minipage}
\caption{Spectrum of fake images}
\label{Fig2}
\end{figure}

In the experiment, CycleGAN models were trained under
two conditions: with checkerboard artifacts (conventional) 
and without checkerboard artifacts (proposed).
Fake images generated by using the models were applied to 
a fake-image detection method\cite{AutoGAN}.
\subsection{EXPERIMENTAL SETUP}
To train models, we used the same generator, discriminator
structures and hyper parameters as those in \cite{CycleGAN}. 
The apple2orange dataset was also used for 
training models and testing.\cite{CycleGAN} 
This dataset consists of apple-images and orange-ones, so models were trained
for designing an unpaired image-to-image transformation such as 
from apples to oranges.
\subsection{EXPERIMENTAL RESULTS}
Figure \ref{Fig1} shows an example of fake images generated by 
using the trained models, where the image in Fig.\ref{Fig1} (a) is 
the input one (real). 
The image in Fig.\ref{Fig1} (b) includes checkerboard artifacts, 
although it is not easy to visually find the artifacts. 
To clearly show the difference between two the fake images, 
the images were represented in the frequency domain as in Fig.\ref{Fig2}, 
where one-dimensional (1D) signals in the horizontal direction were also used
to compute logarithmic spectrums as well as in the paper\cite{AutoGAN}.
From Fig.\ref{Fig2}, the proposed scheme was demonstrated to be effective in
removing checkerboard artifacts even when GANs were applied to 
image to image transformation for generating fake images for the first time.

Next, the effectiveness of the proposed CycleGAN was evaluated in terms of
detection accuracy under the use of a fake-image detection method\cite{AutoGAN}.
Experimental results are shown in Table \ref{Detect}, where w/o artifacts
corresponds to the proposed method, and two accuracy indices: 
ACC and ACC (Fake) in Table \ref{Detect} are defined by

\begin{eqnarray}
ACC = \frac{N_{tn} + N_{tp}}{N_{Qf} + N_{Qr}} \\
ACC(Fake) = \frac{N_{tn}}{N_{Qf}}.
\end{eqnarray}
$N_{Qf}$ is the number of fake query images and $N_{Qr}$ is 
the number of real query ones in the data set. 
$N_{tn}$ also indicates the number of true negatives which are outcomes 
where the model correctly predicts the negative class, and $N_{tp}$ denotes 
the number of true positives which are outcomes which the model correctly
predicts the positive class.
From Table \ref{Detect}, the detection accuracy decreased when using 
the proposed GAN(w/o artifacts).
This is because the detection method assumes that checkerboard artifacts are
included in fake images. 
Accordingly, the proposed CyclicGAN was confirmed to be effective 
as a counter forensics method of fake-image detection.

\begin{table}[htb]
\caption{Comparison with conventional CycleGAN}
\begin{center}
\begin{tabular}{c|cccc}
\hline
\textbf{Dataset}&\multicolumn{2}{|c}{\textbf{w/ artifacts}}&\multicolumn{2}{c}{\textbf{w/o artifacts (proposed)}} \\
\cline{2-5} 
\textbf{} & \textbf{ACC}& \textbf{ACC (Fake)}& \textbf{ACC}& \textbf{ACC (Fake)} \\
\hline
\textbf{apple2orange}& 0.85	&0.92	&0.46	&0.12 \\
\hline
\end{tabular}
\label{Detect}
\end{center}
\end{table}

\section{CONCLUSION}
\label{sec:conclusion}
We proposed a novel CycleGAN without checkerboard artifacts for 
the first time, which allows us to reduce the detection accuracy of 
state-of-the-art fake-image detection methods. 
In the proposed CycleGAN, a fixed convolution layer is inserted into 
not only every upsampling layer but also every downsampling layer in generators.
In the experiment, images generated by the proposed one were applied to 
a fake-image detection method so that
the proposed method was also demonstrated to be effective 
in terms of detection accuracy of fake–images.

\bibliography{report} 

\begin{thebibliography}{10}

\bibitem{Warit_1}
Sirichotedumrong, W., Maekawa, T., Kinoshita, Y., and Kiya, H.,
  ``Privacy-preserving deep neural networks with pixel-based image encryption
  considering data augmentation in the encrypted domain,'' {\em IEEE
  International Conference on Image Processing}  (2019).

\bibitem{Warit_2}
Sirichotedumrong, W. and Kiya, H., ``Grayscale-based block scrambling image
  encryption using ycbcr color space for encryption-then-compression systems,''
  {\em APSIPA Trans. Signal and Information Processing}~{\bf 8},  no.E7 (2019).

\bibitem{Warit_3}
Sirichotedumrong, W., Kinoshita, Y., and Kiya, H., ``Pixel-based image
  encryption without key management for privacy-preserving deep neural
  networks,'' {\em IEEE Access}~{\bf 7},  177844--177855 (2019).

\bibitem{Maung_1}
AprilPyone, M., Kinoshita, Y., and Kiya, H., ``Adversarial robustness by one
  bit double quantization for visual classification,'' {\em IEEE Access}~{\bf
  7},  177932--177943 (2019).

\bibitem{Maung_2}
AprilPyone, M. and Kiya, H., ``Block-wise image transformation with secret key
  for adversarially robust defense,'' {\em arXiv:2010.00801}  (2020).

\bibitem{Maung_3}
AprilPyone, M. and Kiya, H., ``Encryption inspired adversarial defense for
  visual classification,'' {\em IEEE International Conference on Image
  Processing}  (2020).

\bibitem{GAN}
Goodfellow, I.~J., Pouget-Abadie, J., Mirza, M., Xu, B., Warde-Farley, D.,
  Ozair, S., Courville, A., and Bengio, Y., ``Generative adversarial nets,''
  {\em Neural Information Processing Systems}  (2014).

\bibitem{AutoGAN}
Zhang, X., Karaman, S., and Chang, S.-F., ``Detecting and simulating artifacts
  in gan fake images,'' {\em IEEE International Workshop on Information
  Forensics and Security}  (2019).

\bibitem{CNNdetect}
Wang, S.-Y., Wang, O., Zhang, R., Owens, A., and Efros, A.~A., ``Cnn-generated
  images are surprisingly easy to spot... for now,'' {\em Computer Vision and
  Pattern Recognition} ,  8695--8704 (2020).

\bibitem{Kinoshita}
Kinoshita, Y. and Kiya, H., ``Fixed smooth convolutional layer for avoiding
  checkerboard artifacts in cnns,'' {\em IEEE International Conference on
  Acoustics, Speech and Signal Processing} ,  3712--3716 (2020).

\bibitem{Sugawara_1}
Sugawara, Y., Shiota, S., and Kiya, H., ``Checkerboard artifacts free
  convolutional neural networks,'' {\em APSIPA Trans. Signal and Information
  Processing}~{\bf 8},  no.E9 (2019).

\bibitem{Sugawara_2}
Sugawara, Y., Shiota, S., and Kiya, H., ``Super-resolution using convolutional
  neural networks without any checkerboard artifacts,'' {\em IEEE International
  Conference on Image Processing}  (2018).

\bibitem{CycleGAN}
Zhu, J.-Y., Park, T., Isola, P., and Efros, A.~A., ``Unpaired
  image-to-imagetranslation using cycle-consistent adversarial networks,'' {\em
  International Conference of Computer Vision}  (2017).

\bibitem{SpectralNorm}
Miyato, T., Kataoka, T., Koyama, M., and Yoshida, Y., ``Spectral normalization
  for generative adversarial networks,'' {\em International Conference on
  Learning Representations}  (2018).

\end{thebibliography}
\bibliographystyle{spiebib} 

\end{document}